%% file: main.tex
\documentclass{article} 

\input{math_commands.tex}

\usepackage[numbers,square]{natbib}
\usepackage{float}

\usepackage{hyperref}
\usepackage{url}

\usepackage[margin=1in]{geometry}
\usepackage{amsmath,amssymb,bm}
\usepackage{booktabs}
\usepackage{graphicx}
\usepackage{xcolor}
\usepackage[protrusion=false]{microtype}
\usepackage{enumitem}

\newcommand{\ORBIT}{\textsc{ORBIT}}
\newcommand{\ORBITFMIB}{\textsc{ORBIT-FMIB}}

\title{
\textbf{ORBIT-FMIB: Tracking Order-Resolved Epistatic Information Through ESM-2}
}

\author{
Maryam Rahimimovassagh \\
University of Central Florida
\and
Waldemar Karwowski \\
University of Central Florida
\and
Ivan Garibay \\
University of Central Florida
\and
Niloofar Yousefi \\
University of Central Florida
}

\date{}

\input{orbit_result_macros.tex}

\begin{document}

\maketitle

\begin{abstract}
\normalsize

Protein foundation models support mutation-effect, fitness, and structural
prediction, but predictive performance alone does not reveal which forms of
biological interaction information remain accessible through model depth.
Here, we ask whether ESM-2 retains higher-order epistatic information as
strongly as first- and second-order information across its representation
hierarchy.

We introduce \ORBITFMIB{}, \textbf{Order-Resolved Benchmarking of Interaction
Transformations for Foundation-Model Information Bottlenecks}, a diagnostic
framework that combines Walsh-based interaction decomposition with
subset-conditioned neural dependence estimation. The method traces
experimentally aligned first-, second-, third-, and fourth-order epistatic
components through model depth while estimating each positional subset
independently before aggregation by interaction order. We first validate the
complete pipeline on \ORBIT-Synth, where interaction orders and
information-loss events are known, and then apply it to the dense four-site
GB1 fitness landscape using frozen ESM-2 representations.

ESM-2 exhibited an order-dependent retention pattern in an initial
production run. Peak-to-final retention was 0.960 and 0.971 for first-
and second-order interactions, compared with 0.910 and 0.808 for third-
and fourth-order interactions, respectively. Mean higher-order retention
was lower than mean lower-order retention (0.859 versus 0.966;
$\Delta_{\mathrm{HO-LO}}=-0.107$). A whole-identity deletion sensitivity
analysis preserved the same direction ($\Delta_{\mathrm{HO-LO}}=-0.184$).

An independent replication of the complete measurement grid, using identical code, data, and critic seeds, substantially reduced this contrast ($\Delta_{\mathrm{HO-LO}}=-0.017$), and its sign was not
stable across otherwise-defensible evaluation-pairing choices applied
to the same trained critics (ranging $-0.011$ to $+0.015$). We therefore
do not currently have robust evidence that ESM-2 selectively loses
higher-order epistatic information, nor evidence that retention is
equal across orders; the directional question remains open. The
fixed-subset measurement protocol itself including its documented
removal of a positional-subset shortcut in pooled critics is
unaffected by this finding. \ORBITFMIB{} is offered as a diagnostic
framework for probing how complex biological interaction structure is
transformed within protein foundation models, and this study's own
replication result illustrates why such probing requires
adequately-powered reproducibility checks before its output is treated
as a biological finding.

\end{abstract}

\clearpage
\section{Introduction}

Protein sequence--function relationships are inherently non-additive.
The effect of an amino-acid substitution can depend on residues elsewhere in
the sequence, producing epistasis. When a joint mutational effect cannot be
explained by individual or lower-order contributions, the residual component
constitutes higher-order epistasis \cite{zhou2022higher}. Such interactions
are biologically important because protein function emerges from
three-dimensional structures in which residues that are distant in sequence
can become physically and functionally coupled.

Beyond its structural interpretation, understanding epistatic structure has
direct practical consequences. In protein engineering and directed evolution,
non-additive mutational effects mean that combinatorial libraries and
ML-guided search strategies must account for interaction structure rather
than assuming individually-beneficial mutations combine additively
\cite{miton2016mutational}. The degree and pattern of epistasis
also shapes the ruggedness of a fitness landscape, which in turn influences
how readily resistance mutations emerge and combine under selective pressure,
with direct relevance to predicting and anticipating drug-resistance
evolution in pathogens \cite{palmer2015delayed}. In clinical genetics, whether co-occurring variants combine
additively or epistatically affects penetrance and severity prediction in
polygenic conditions \cite{wei2025investigating}.
Antibody-antigen binding interfaces frequently exhibit epistatic coupling
among contact residues, making this structure relevant to predicting viral
escape mutations and designing broadly neutralizing antibodies
\cite{witte2023epistasis}. If protein language models
such as ESM-2 encode higher-order epistatic structure despite never being
trained on fitness measurements, this would offer a transferable,
experiment-free proxy for these applications across new proteins; if they do
not, applications relying on such models to implicitly capture higher-order
interaction effects would need to be reassessed accordingly.

Epistasis can therefore reflect underlying biophysical constraints, including
folding stability, steric compatibility, residue packing, binding, and
conformational coupling. In GB1, mutational studies have linked epistatic
effects to structural stability and interacting residue networks
\cite{olson2014biophysical,wu2016adaptation}. More broadly, epistatic
measurements from deep mutational scans can identify residue contacts and
support reconstruction of three-dimensional protein structure
\cite{rollins2019inferring,schmiedel2019determining}. In the four-site GB1
landscape considered here, previous work has specifically connected strong
interactions involving sites 41 and 54 to steric effects in the folded
structure \cite{wu2016adaptation,zhou2020minimum}.

This makes epistatic information a biologically meaningful probe of learned
representations. A representation may preserve individual residue effects
while making joint multi-residue dependencies less accessible, potentially
weakening information about the cooperative constraints through which
sequence determines structure and function.

A related distinction arises in gene regulatory network inference, where
strong pairwise edge rankings do not necessarily imply recovery of complete
cooperative regulator sets. Residual set modeling and the BRIDGE framework
address this distinction through explicit set-level scoring and recovery
diagnostics \citep{rahimimovassagh2026beyondedge,rahimimovassagh2026bridge}.
These studies motivate evaluating joint biological interactions explicitly,
beyond aggregate predictive performance.

Protein language models provide a natural setting in which to ask whether
such information is preserved. Sequence-trained models have been shown to
encode evolutionary, functional, and structural properties
\cite{rives2021biological,rao2021unsupervised,pmlr-v139-rao21a}.
Transformer attention can recover residue contacts without direct structural
supervision \cite{rao2021unsupervised}, while protein language-model
representations support mutation-effect prediction
\cite{meier2021language,notin2023proteingym} and atomic-level structure
prediction \cite{lin2023evolutionary}. Related sequence models also show that
capturing dependencies among positions improves prediction of mutational
effects \cite{hopf2017mutation,riesselman2018deep}. Yet successful prediction
or the presence of structural information somewhere in a model does not imply
that all biologically relevant interaction structure remains equally
accessible throughout its representation hierarchy
\cite{zhang2024motifs}.

Recent work suggests that interaction order itself is a meaningful axis for
analyzing Transformers. Rende et al.\ showed that lower-order dependencies are
learned earlier than higher-order dependencies during Transformer training
\cite{rende2024distributional}. In protein language models, Tsui and
Aghazadeh recovered higher-order mutational interactions from ESM-2 outputs,
including GB1, using sparse Fourier analysis
\cite{tsui2024recovering}. However, these studies do not determine how
experimentally measured interaction orders evolve through the intermediate
hidden representations of a pretrained protein model.

Complementary work introduced ORBIT, a diagnostic framework that separates
interaction presence, representation accessibility, and functional recovery
using synthetic landscapes and task-trained models of GB1 fitness
\cite{rahimimovassagh2026beyondtokens}. The present work builds on this
order-resolved perspective to investigate frozen protein foundation-model
representations. Specifically, ORBIT-FMIB combines interaction decomposition
with subset-conditioned neural dependence estimation to examine how
experimentally aligned epistatic components are retained across model depth.

We address this gap through an interaction-order-resolved view of information
flow. Classical information-bottleneck analyses ask how representations
trade input information against target-relevant information
\cite{tishby2000information}, and FlowNIB extends this view to layer-wise
mutual information in language models \cite{kowsher2026flownib}, but a global
quantity such as $I(Z_\ell;Y)$ cannot distinguish whether retained fitness
information reflects individual residue effects, pairwise interactions, or
genuine third- and fourth-order epistasis.

We therefore ask:

\begin{quote}
\textbf{Does ESM-2 retain third- and fourth-order epistatic information as
strongly as first- and second-order information across its representation
hierarchy?}
\end{quote}

We introduce ORBIT-FMIB, a diagnostic framework for tracing experimentally aligned epistatic components through frozen protein-model representations. The framework applies a common Walsh decomposition to experimental fitness and hidden representations, then estimates their dependence separately for each fixed positional subset. This subset-specific measurement is essential: our adversarial controls reveal that pooling subsets can produce apparent dependence driven by subset identity, even after the matched cube-level relationship is disrupted. Synthetic validation tests whether the corrected protocol recovers known interaction orders and imposed changes in dependence before application to GB1.

Our contribution has two parts: a measurement protocol that addresses positional-subset confounding, and an empirical characterization of first- through fourth-order epistatic dependence across ESM-2 depth. In an initial production run on the four-site GB1 landscape, higher-order components showed lower aggregate peak-to-final retention, with substantial variation among third-order subsets and the strongest reduction in the sole fourth-order component. A subsequent independent replication of the complete measurement grid (Section~\ref{sec:replication}) substantially narrowed this contrast and found its sign unstable across evaluation-pairing choices applied to the same trained critics, so this pattern should not be read as an established property of ESM-2. The measurement protocol itself -- including its validated removal of the positional-subset shortcut -- is unaffected by this finding. These findings characterize what is and is not currently supported under the validated critic protocol, and motivate both testing which representation stages best preserve experimentally relevant multi-residue effects and characterizing the estimator's own run-to-run variance, while leaving structural consequences, causal model use, and generalization across proteins and models as open questions.


\begin{figure*}[!t]
    \centering
    \includegraphics[
        width=\textwidth,
        keepaspectratio
    ]{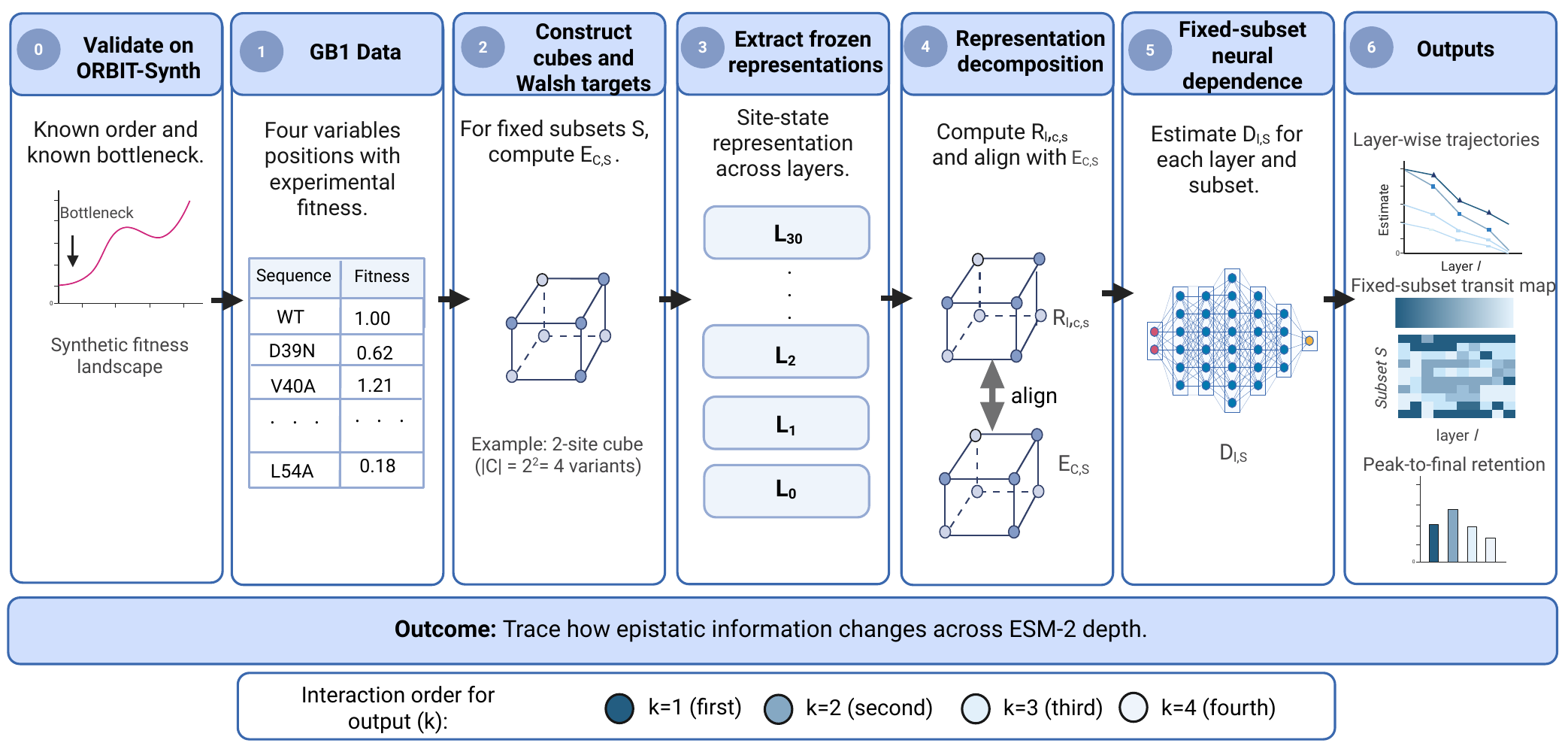}
    \caption{
    \textbf{Overview of the ORBIT-FMIB pipeline.}
    The measurement framework is first validated on \ORBIT-Synth{}, where
    interaction orders and imposed information-loss events are known, and is
    then applied to the four-site GB1 fitness landscape using frozen ESM-2
    representations. Complete WT-anchored mutational cubes are decomposed
    using a common Walsh basis to obtain experimental epistatic coefficients
    $E_{C,S}$ and representation-level interaction components
    $R_{\ell,C,S}$. Neural dependence $D_{\ell,S}$ is estimated independently
    for each fixed positional subset across model depth, yielding layerwise
    trajectories and peak-to-final retention summaries.
    }
    \label{fig:orbit_pipeline}
\end{figure*}

\section{Method and Experimental Setup}

\ORBITFMIB{} traces experimentally aligned epistatic information through
model depth for fixed positional interaction subsets. For each mutational
cube $C$, representation stage $\ell$, and subset $S$, the same Walsh basis
is applied to the hidden representation and experimental fitness landscape,
yielding $R_{\ell,C,S}$ and $E_{C,S}$. A separate DV/MINE critic is fit for
each fixed layer--subset pair,

\begin{equation}
D_{\ell,S}
=
\widehat{I}_{\mathrm{DV/MINE}}
\left(R_{\ell,C,S};E_{C,S}\right).
\end{equation}

Holding $S$ fixed prevents the positional-subset shortcut identified during
estimator stress testing. Full mathematical definitions, estimator details,
calibration, and null controls are provided in
Appendix~\ref{app:full_method}.

\paragraph{Synthetic validation.}
Before biological interpretation, ORBIT-FMIB is validated on \ORBIT-Synth,
where interaction orders and information-loss events are known. The pipeline
passed pre-specified recovery, null, retention, and adversarial-pattern checks.
Because absolute mutual-information magnitude was not calibrated at the
biological representation dimension, biological scores are interpreted
comparatively rather than as exact mutual information in nats
(Appendix~\ref{app:estimator_validation}).

\paragraph{GB1 and ESM-2.}
We analyze the four-site GB1 landscape at positions 39, 40, 41, and 54
\cite{wu2016adaptation}, retaining 8,210 complete WT-anchored cubes.
Frozen representations from \texttt{esm2\_t30\_150M\_UR50D}
\cite{lin2023evolutionary} are extracted across 31 stages, with the four site
vectors concatenated into a 2,560-dimensional representation. Cubes are
partitioned into 7,352 training, 245 validation, and 613 test observations.
Full construction, extraction, indexing, and split details are provided in
the Appendix.

This partition separates cubes for critic training, validation, and evaluation, but WT-anchored cubes can share constituent sequences and amino-acid identities across partitions. It therefore serves as an IID cube-level estimator split and does not establish generalization to unseen protein variants or independent biological sampling. Held-out test epistatic coefficients are excluded from critic optimization, standardization, and model selection. The resulting dependence trajectories and retention contrasts are interpreted descriptively within this overlapping cube population.

\paragraph{Primary analysis.}
Each of the 15 nonempty positional subsets yields an independently estimated
layerwise trajectory. Retention is computed within subset before aggregation
by interaction order, with the primary descriptive contrast

\begin{equation}
\Delta_{\mathrm{HO-LO}}
=
\frac{\mathrm{Retain}_3+\mathrm{Retain}_4}{2}
-
\frac{\mathrm{Retain}_1+\mathrm{Retain}_2}{2}.
\end{equation}

A negative value indicates lower higher-order retention. No formal $p$-value
is assigned; critic seeds assess estimator stability rather than biological
sampling uncertainty. Full statistical and sensitivity procedures are given
in Appendix~\ref{app:statistics}.

\section{Results}

\subsection{ORBIT-Synth Validates Order-Resolved Information Tracking}

Before biological interpretation, we evaluated ORBIT-FMIB on ORBIT-Synth,
where interaction order and imposed information-loss locations are known by
construction. The benchmark included first-, second-, third-, and fourth-order
signals together with controlled representation trajectories in which selected
interaction components were attenuated at known stages.

The complete pipeline recovered the intended interaction-order structure and
correctly localized the imposed order-specific information-loss events. In the
staggered-bottleneck benchmark, recovered loss stages matched the known
synthetic construction for all four interaction orders.

At the exact 2,560-dimensional representation size used in the biological
analysis, the neural estimator preserved the expected qualitative ordering of
dependence strength but underestimated moderate and strong analytic mutual
information. Biological scores are therefore interpreted as comparative neural
dependence rather than calibrated mutual information in nats. Additional relative-retention and hypothesis-adversarial tests confirmed the
expected direction for higher-order-loss and higher-order-advantage
scenarios; the flat-retention criterion's statistical power is addressed in
Limitations.

\subsection{Fixed-Subset Conditioning Removes Positional Subset-Identity Confounding}

The initial pooled estimator failed an adversarial cube-derangement
control, revealing positional subset identity as a potential shortcut.
Different positional subsets within the same interaction order
exhibited systematically different epistatic-target distributions, allowing a pooled critic to retain apparent
dependence even after the matched cube-level relationship was disrupted.

We therefore reformulated the analysis so that each neural critic was
conditioned on a single fixed positional subset $S$. Under this correction,
cube-deranged null scores collapsed near the estimator floor, while matched
finite-sample synthetic controls retained detectable dependence. All
biological results below therefore use independently estimated fixed-subset
critics, with interaction-order summaries formed only after subset-level
measurement.

\subsection{Epistatic Information Follows Order-Dependent Trajectories Through ESM-2}

We traced all 15 nonempty positional subsets of the four-site GB1 landscape
across the 31 ESM-2 representation stages, $\ell=0,\ldots,30$. Each subset was
estimated independently, producing one layerwise neural-dependence trajectory
before interaction-order aggregation.

\begin{figure*}[!t]
    \centering
    \includegraphics[width=0.49\textwidth]{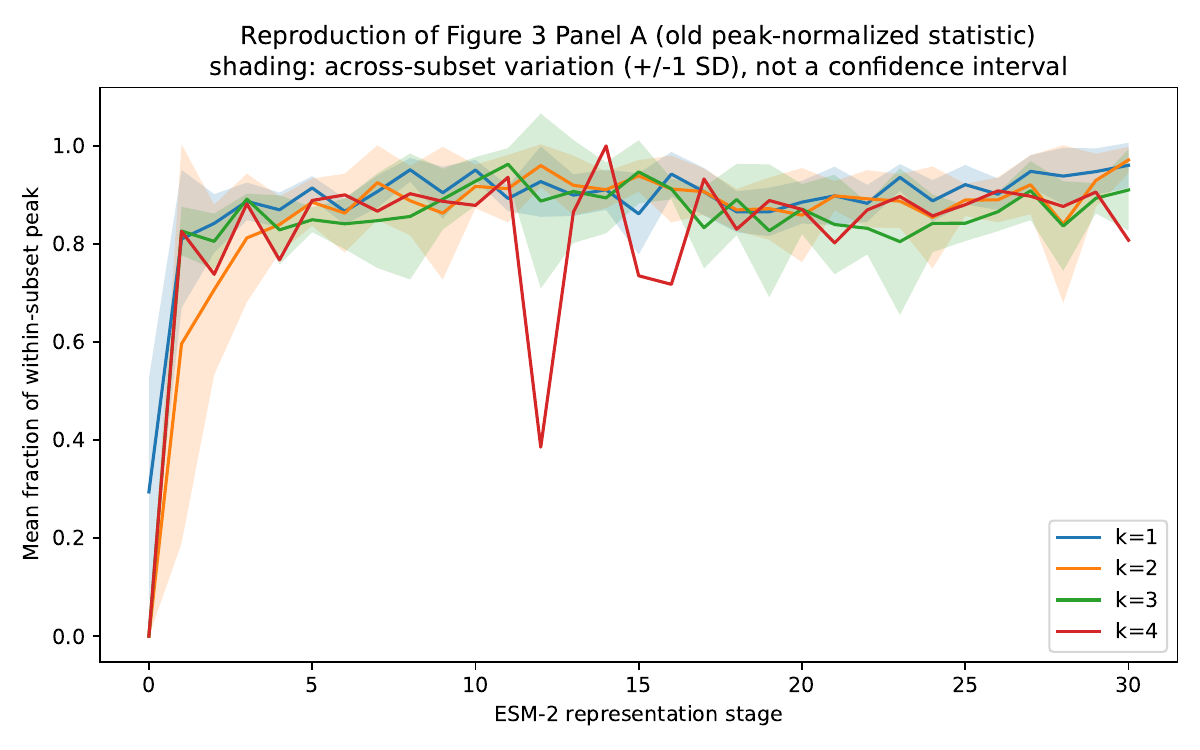}
    \hfill
    \includegraphics[width=0.49\textwidth]{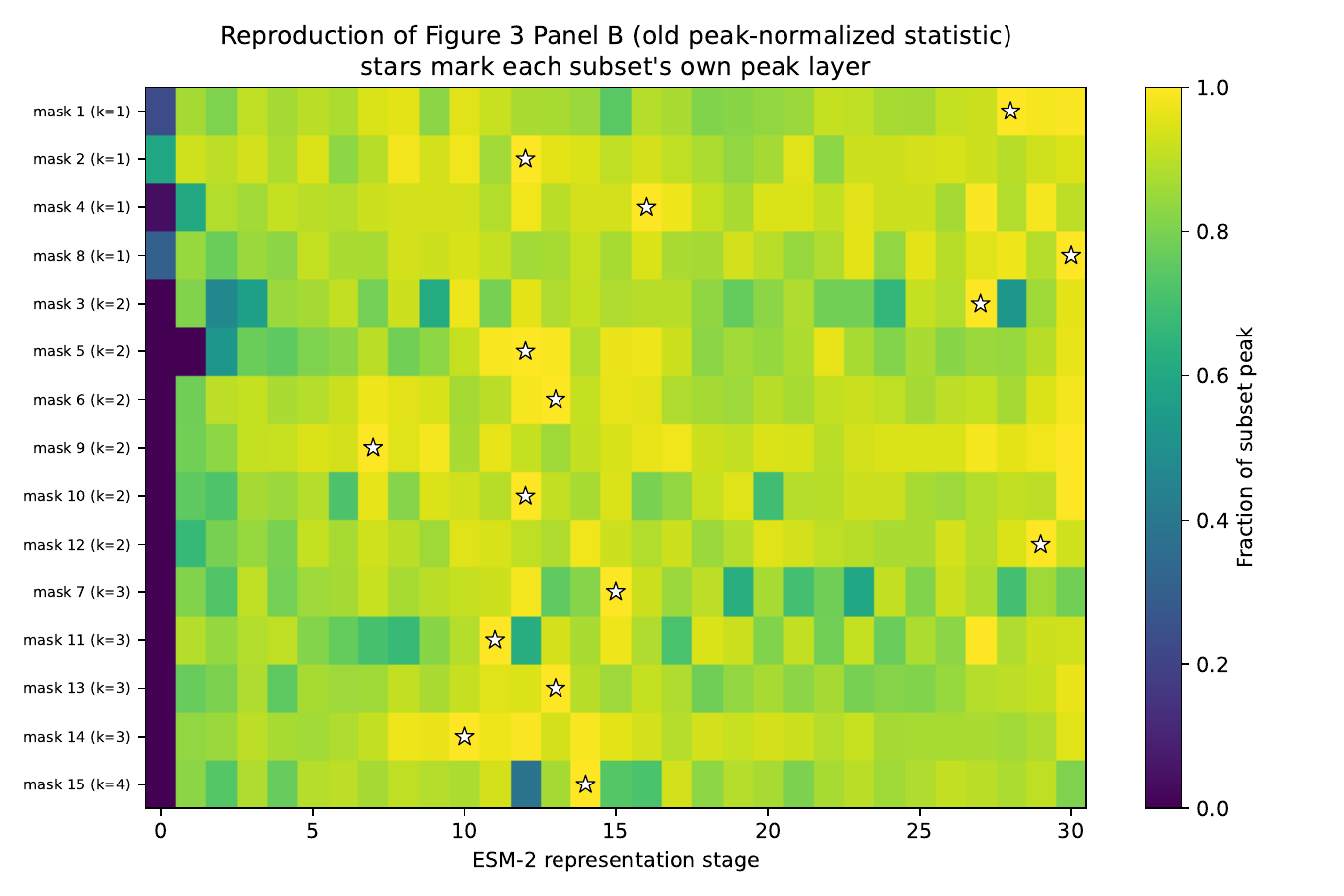}
    \caption{
    \textbf{Order-resolved epistatic information dynamics across ESM-2 depth
    (production run).}
    \textbf{A,} Order-level trajectories after normalization of each fixed
    subset to its own peak; shading summarizes across-subset variation
    ($\pm 1$ SD) and is not a confidence interval.
    \textbf{B,} Peak-normalized trajectories for all 15 fixed positional
    subsets across L0--L30; stars mark each subset's own peak layer.
    Neural dependence scores are comparative and are not interpreted as
    calibrated mutual information in nats.
    }
    \label{fig:gb1_layerwise_dynamics}
\end{figure*}

For visualization, each fixed-subset trajectory was normalized to its own
maximum and subsequently summarized by interaction order
(Fig.~\ref{fig:gb1_layerwise_dynamics}). This preserves within-subset dynamics
while avoiding direct comparison of raw neural-dependence magnitudes across
different subsets.

The resulting trajectories differed by interaction order. The descriptive
first- and second-order curves reached their maxima at L30, with final
peak-normalized values of 0.960 and 0.971. In contrast, the third-order curve
peaked at L11 and declined to 0.910, whereas the fourth-order trajectory peaked
at L14 and declined to 0.808 by L30. Thus, lower-order information remained
near its strongest observed alignment through the final ESM-2 stage, while
higher-order trajectories showed greater peak-to-final attenuation. These
trajectories are from a single production run; Section~\ref{sec:replication}
reports that an independent replication of the full grid did not reproduce
this pattern at comparable magnitude and found its sign unstable across
evaluation-pairing choices.

\subsection{Fixed-Subset Trajectories Reveal Within-Order Heterogeneity}

Interaction order did not fully determine individual subset trajectories.
First-order subset peaks ranged from L12 to L30, second-order peaks from L7 to
L29, and third-order peaks from L10 to L15.

Third-order peak-to-final retention also varied substantially:
$\{39,40,41\}$ retained 0.788 of its peak,
$\{39,40,54\}$ retained 0.929,
$\{39,41,54\}$ retained 0.970, and
$\{40,41,54\}$ retained 0.953. The sole fourth-order subset
$\{39,40,41,54\}$ retained 0.808. Thus, interaction order captures a broad
retention pattern, but the fixed-subset trajectories remain the primary
measurement objects. These per-subset values are from the single production
run; per-subset values under independent replication are given in
Appendix~\ref{app:replication}.

\subsection{Higher-Order Epistatic Information Shows Lower Peak-to-Final Retention}

Order-level peak-to-final retention was
\OrbitPrimaryRetainOne{} for $k=1$,
\OrbitPrimaryRetainTwo{} for $k=2$,
\OrbitPrimaryRetainThree{} for $k=3$, and
\OrbitPrimaryRetainFour{} for $k=4$.
Mean lower-order retention was \OrbitPrimaryLowOrderRetention{}, compared with
\OrbitPrimaryHighOrderRetention{} for the higher-order group, yielding
$\Delta_{\mathrm{HO-LO}}=\OrbitPrimaryDelta{}$.

The negative contrast indicates lower aggregate peak-to-final retention for
third- and fourth-order than for first- and second-order epistatic information.
This should not be interpreted as uniform higher-order loss: three of the four
third-order subsets retained more than 0.92 of their peaks, while one showed
substantially greater attenuation. The strongest peak-to-final reduction was
observed for the single fourth-order component. These values are from a
single production run; Section~\ref{sec:replication} reports that this
contrast did not reproduce at comparable magnitude, and its sign was not
stable, under independent replication of the complete measurement grid.
\subsection{Whole-Identity Deletion Preserves the Direction of the Retention Pattern}

We next tested whether the primary result was dominated by one repeatedly
represented amino-acid identity. A dependence-aware sensitivity analysis
removed every eligible mutational cube containing the alternative identity
$41=L$.

The deletion retained
\OrbitSensitivityTotalN{} complete cubes, partitioned into
\OrbitSensitivityTrainN{} training,
\OrbitSensitivityValN{} validation, and
\OrbitSensitivityTestN{} test cubes.

Following deletion, order-level retention was
\OrbitSensitivityRetainOne{},
\OrbitSensitivityRetainTwo{},
\OrbitSensitivityRetainThree{}, and
\OrbitSensitivityRetainFour{}
for interaction orders $k=1,2,3,4$, respectively
(Table~\ref{tab:order_retention_sensitivity}).

The lower-order retention summary was

\begin{equation}
\mathrm{LO}_{\mathrm{sens}}
=
\OrbitSensitivityLowOrderRetention,
\end{equation}

whereas higher-order retention was

\begin{equation}
\mathrm{HO}_{\mathrm{sens}}
=
\OrbitSensitivityHighOrderRetention.
\end{equation}

The corresponding contrast was

\begin{equation}
\Delta_{\mathrm{HO-LO}}^{\mathrm{sens}}
=
\OrbitSensitivityDelta.
\end{equation}

The direction of the primary higher-versus-lower-order separation therefore
persisted after removal of all cubes containing $41=L$ and became larger in
magnitude. The sensitivity analysis is descriptive: it changes the underlying
cube population and is not treated as an independent biological replicate or
as a formal statistical test. Like the primary result, this sensitivity check
was run once, on the same single production critics; Section~\ref{sec:replication}
reports that the underlying contrast it is checking did not itself reproduce
under independent replication.

\input{tables/table_order_retention_primary_sensitivity.tex}

\subsection{Independent Replication and Evaluation-Pairing Sensitivity}
\label{sec:replication}

The results above come from a single completed production run. To test
whether the order-selective contrast reproduces under an independent run
of the same measurement protocol, we retrained the complete
31-layer $\times$ 15-subset $\times$ 3-critic-seed grid (1,395 fits)
from scratch, using identical code, identical GB1/ESM-2 data (matching
upstream provenance hashes for every prior pipeline stage), the same
deterministic IID cube partition, and the same critic seeds (11, 23, 37)
as the original production run. Training used an NVIDIA H100 GPU. The original production run's manifest records only \texttt{device\_type: cuda}; its specific GPU model was not recorded, so this cannot be confirmed as a hardware match.

This replication is not a reproduction of the original checkpoints.
The mechanism underlying this discrepancy remains unresolved. A targeted execution-environment diagnostic is reported in Appendix~\ref{app:execution_environment}. Only 318 of the
1,395 replicated fits (23\%) selected the same best-validation training
step as the original run, and the mean absolute difference in held-out
score from the original production values was 0.081 nats (maximum 0.688
nats for a single layer--subset--seed cell). We interpret this as
evidence that the two runs produced genuinely different trained
critics at most layer--subset cells, not merely noisy re-evaluations of
the same critic.

For every replicated critic, we additionally computed three held-out
evaluations from the same restored checkpoint and the same standardized
test tensors, motivated by the evaluation-pairing concern already noted
in Appendix~\ref{app:estimator_validation}: (i) the original cyclic-shift
pairing used throughout the primary analysis above; (ii) the same cyclic
shifts applied after jointly permuting the row order of the
representation and target vectors together, which preserves true pairs
while testing sensitivity to row order; and (iii) negative pairs drawn
by full random permutation rather than fixed shifts. All three read out
the same trained critic; none involve additional retraining.

Using the original cyclic-shift evaluation, the replication yielded
$\Delta_{\mathrm{HO-LO}}=-0.017$, an 84\% reduction in magnitude from the
original production value of $-0.107$. The two alternative evaluation
pairings applied to the same replicated critics gave
$\Delta_{\mathrm{HO-LO}}=-0.011$ (joint-row permutation) and $+0.015$
(random permutation), respectively -- see the replication columns of
Table~\ref{tab:order_retention_sensitivity} above for the full
order-level comparison alongside the original production and
whole-identity-deletion values.

The sign of the order-selective contrast is therefore not stable across
otherwise-defensible evaluation choices applied to the same trained
critics, and its magnitude is not stable across independent training
runs under the original evaluation choice. We do not interpret this as
evidence that higher-order information is retained equally to
lower-order information in ESM-2: the replication and the two
alternative pairings are themselves single runs, not a characterization
of the estimator's sampling distribution, and no formal uncertainty
quantification has been performed. What this result does establish is
that the originally reported $\Delta_{\mathrm{HO-LO}}=-0.107$ is not a
stable, reproducible summary of this measurement protocol under
nominally identical code, data, and critic seeds (the original run's execution hardware was not recorded; Appendix P). Characterizing the estimator's true
run-to-run variance -- via multiple independent replications, not a
single alternative run -- is identified in Future Work as necessary
before any directional claim about order-dependent retention in ESM-2
can be supported. Full per-subset results under all three evaluators are
given in Appendix~\ref{app:replication}.

\section{Discussion}

ORBIT-FMIB's fixed-subset measurement protocol -- including its
validated removal of a positional-subset shortcut in pooled critics --
is itself reproducible; that methodological contribution is unaffected
by the finding below. The biological question it was applied to is a
different matter. An initial production run suggested an order-dependent
retention pattern in ESM-2: first- and second-order epistatic
information remained close to its strongest observed alignment at the
final representation, whereas third- and especially fourth-order
information showed greater peak-to-final attenuation. An independent
replication of the complete measurement grid
(Section~\ref{sec:replication}) did not reproduce this contrast at
comparable magnitude, and found its sign unstable under alternative,
equally defensible evaluation-pairing choices applied to the same
trained critics. We therefore do not treat selective higher-order
information loss in ESM-2 as an established finding of this study, and
we equally do not have evidence that retention is uniform across order;
the directional question is open.

This matters beyond the specific numbers because of what it implies
about measurement, not just about ESM-2. The layerwise view remains
methodologically important -- endpoint analysis alone cannot distinguish
information that is weak throughout the network from information that
becomes strongly represented at intermediate depth and later becomes
less detectable -- but the present replication result shows that this
distinction, and the retention values used to draw it, are themselves
sensitive to which specific trained critic and which evaluation pairing
is used. Critic-seed variation was already known to shift the apparent
peak layer; the replication result extends this to the retention
statistic itself.

We had originally hypothesized that reduced accessibility of
higher-order epistatic information at later ESM-2 layers might reflect
genuine attenuation of cooperative structural and functional
constraints among residues. That hypothesis presupposes a stable
underlying measurement to interpret, which we do not currently have.
Rather than a specific representational asymmetry to be explained
biologically, the present study's central finding is a characterization
of how unstable this particular measurement protocol's output is across
independent runs and evaluation choices -- a prerequisite that needs to
be resolved before the biological hypothesis can be tested at all.

\section{Limitations}

The present conclusions are intentionally narrow. The biological analysis is
restricted to the four-site GB1 landscape, and the fourth-order category
contains only one positional subset, so the result does not establish a
general property of fourth-order protein interactions. The neural dependence
score is comparative rather than a calibrated estimate of absolute mutual
information, and peak timing is descriptive rather than a mechanistically
identified transition point. WT-anchored cubes can share sequences and amino-acid
identities, so the deterministic IID partition is an estimator split rather
than a claim of independent biological sampling. ESM-2 may also have encountered
Protein G or related homologs during pretraining, limiting claims of
sequence-family novelty.

Several validation gaps also limit how much the observed higher-order/lower-order
contrast can be attributed to biology rather than to the estimator. The flat-pattern acceptance test used in the synthetic adversarial suite
was originally implemented as a permutation test whose minimum attainable
two-sided $p$-value is $1/3$ given only four order labels, so it could
not fail at the $\alpha=0.05$ threshold regardless of the estimator's
true selectivity. This was corrected using a properly-powered
multi-replicate test, which found no detectable deviation from flatness
under either tested scenario (Appendix~H.4). The null-derangement
diagnostics reported in the appendix were originally computed under the
earlier amino-acid-identity-strict estimator split rather than the
deterministic IID split used for the production GB1 scores; this has
since been re-run under the IID split, and every downstream statistic
was unchanged (Appendix~H.6). Held-out DV/MINE evaluation also pairs representation and target vectors
using fixed cyclic shifts of sorted-index batches rather than full reshuffling, which
may leave residual structure between nearby observations. We tested this last concern
directly (Section~\ref{sec:replication}): under the same independently retrained
critics, alternative evaluation pairings shifted $\Delta_{\mathrm{HO-LO}}$ from
$-0.017$ (original pairing) to values ranging $-0.011$ to $+0.015$, and the
independent replication under the original pairing had already reduced the originally
reported $-0.107$ to $-0.017$. An estimator and/or training-run contribution to the
reported order contrast is therefore confirmed as material -- not merely
unexcluded -- and we do not currently have robust evidence for the original
directional claim in either direction.

Most consequentially, this study's own replication (Section~\ref{sec:replication},
Appendix~\ref{app:replication}) shows that the primary result does not reproduce
at comparable magnitude or sign under an independently retrained critic grid. We
have one replication and three evaluation pairings, not a characterization of the
estimator's sampling distribution; we cannot currently say whether the original
production run or the replication is closer to the estimator's typical behavior,
only that they differ substantially. The replication cannot be verified as hardware-matched: the original run's GPU model was not recorded, and a single-cell diagnostic (Appendix~\ref{app:execution_environment}) shows a V100 and an H100 giving different scores for identical inputs, each reproducing one of the two runs' values for that cell. We therefore cannot attribute the production-versus-replication difference to the training procedure alone. Establishing the estimator's true run-to-run
variance, via multiple independent replications rather than a single alternative
run, is a precondition for any future directional claim about order-dependent
retention using this protocol, and is treated as such in Future Work.

Finally, the biological result is specific to ESM-2. We established frozen
representation extraction for ESM-1v and ProtT5, but their corresponding
high-dimensional neural dependence estimators did not satisfy the same
pre-specified calibration requirements; we therefore excluded cross-model
retention comparisons. Broader claims require independently validated
estimators across additional protein foundation models and landscapes.

\section{Future Work}

The most immediate priority, established by this study's own replication
result, is characterizing the estimator's true run-to-run variance before
any further directional claim is attempted with this protocol. This
requires multiple independent replications of the full measurement grid
(not the single alternative run reported here), on both frozen ESM-2
representations and, as a control, on a raw one-hot mutation encoding
that trivially contains the full Walsh signal by construction -- a
sharp test of whether the observed instability lives in the estimator
and training procedure or specifically in ESM-2's representations.
Diagnosing whether the instability reflects finite-sample variance in a
2,560-dimensional DV/MINE critic, optimization sensitivity to
initialization, or something specific to the cyclic-shift evaluation
pairing is a prerequisite for interpreting any future retention
comparison from this pipeline, including the questions below.

A primary next step, once the estimator's behavior is adequately
characterized, is to determine whether any order-dependent retention
pattern in ESM-2 generalizes across protein foundation models.
Applying the same fixed-subset protocol to additional architectures, including
ESM-1v and ProtT5 once their estimator regimes are independently validated,
would distinguish model-specific behavior from broader representation-learning
effects.

ORBIT-FMIB should also be extended beyond the four-site GB1 landscape.
Larger combinatorial and perturbation datasets would enable higher-order
retention to be studied across more positional subsets and structural
contexts. Structural grounding, alternative readouts, and causal
interventions could then test whether attenuated higher-order information
remains accessible and whether the native model actually uses it -- but
only once the underlying measurement is known to be reproducible.

Methodologically, better-calibrated high-dimensional dependence estimators,
lower-variance alternatives such as ridge-regression decodability probing or
sparse Walsh-coefficient recovery, and dependence-aware resampling could
support more direct cross-representation comparisons and formal uncertainty
quantification. If an order-dependent retention pattern proves reproducible
across multiple independent replications, models, and landscapes, ORBIT-FMIB
could then serve as an evaluation framework for architectures designed to
preserve higher-order biological interaction structure.

\section{Conclusion}

We introduced \ORBITFMIB{}, an order-resolved framework for tracing
experimentally aligned epistatic information through protein
foundation-model representations. Combining Walsh decomposition with
fixed-subset neural dependence estimation allows first- through fourth-order
interaction components to be followed separately through model depth while
avoiding positional subset-identity shortcuts; this measurement protocol
is validated and reproducible.

Applied to the four-site GB1 landscape and frozen ESM-2 representations,
an initial production run of ORBIT-FMIB suggested lower aggregate
peak-to-final retention for higher-order than for lower-order epistatic
information, with the strongest reduction in the fourth-order component.
A subsequent independent replication of the complete measurement grid did
not reproduce this contrast at comparable magnitude, and found its sign
unstable under alternative, equally defensible evaluation-pairing choices
applied to the same trained critics. We therefore do not currently have
robust evidence that ESM-2 selectively loses higher-order epistatic
information, nor evidence that retention is equal across orders; the
directional question raised in our Introduction remains open. This
outcome is itself informative: it demonstrates that retention estimates
from a 2,560-dimensional DV/MINE critic, evaluated with a single training
run and a single evaluation pairing, are not a reliable basis for a
biological claim on their own. ORBIT-FMIB's fixed-subset measurement
protocol, and the reproducibility characterization reported here, are
offered as a framework and a cautionary methodological finding,
respectively, for future work seeking to measure how biological
interaction structure is retained and transformed within protein
foundation models.

\paragraph{Code availability.} Code, configuration files and run manifests for all analyses in this paper are available at \url{https://github.com/maryam-rahimi/ORBIT-FMIB}. Appendices A--P give the full methods, validation and replication details.

%

\bibliographystyle{apalike}
\bibliography{references}

\clearpage
\appendix

\section{Complete ORBIT-FMIB Method and Validation}
\label{app:full_method}

\section{Scope and Contributions}

The study is intentionally restricted to the minimum set of experiments
required to answer the central question.

\paragraph{Biological landscape.}
We use the dense four-site GB1 combinatorial fitness landscape as the sole
biological test system.

\paragraph{Foundation model.}
The primary model is a frozen ESM-2 checkpoint. Model parameters remain fixed
throughout the analysis, and experimentally measured GB1 fitness values are
used only for downstream evaluation of representation--epistasis alignment.

\paragraph{Interaction orders.}
The analysis resolves first-, second-, third-, and fourth-order epistatic
components,

\begin{equation}
k \in \{1,2,3,4\}.
\end{equation}

\paragraph{Primary measurement.}
Dependence is estimated independently for each fixed positional interaction
subset $S$ across model depth. Peak-to-final retention is first computed within
each subset trajectory and only then summarized according to interaction order
$k=|S|$.

\paragraph{Primary comparison.}
The main biological comparison evaluates whether higher-order interactions
($k=3,4$) show lower peak-to-final retention than lower-order interactions
($k=1,2$).

The principal contributions of this study are:

\begin{enumerate}
    \item We formulate protein foundation-model representation analysis as an
    interaction-order-resolved information-flow problem, allowing first-,
    second-, third-, and fourth-order epistatic components to be traced
    separately through model depth.

    \item We develop and validate a subset-conditioned neural-dependence
    pipeline that combines Walsh decomposition with layerwise representation
    analysis while avoiding positional-subset shortcut effects.

    \item We validate the measurement pipeline on controlled synthetic
    representations with known interaction orders and known information-loss
    locations before applying it to biological data.

    \item We apply the validated framework to the GB1 fitness landscape and a
    frozen ESM-2 model. An initial production run showed lower aggregate
    peak-to-final retention for higher-order than for lower-order epistatic
    information, with the largest reduction at the fourth-order component;
    Contribution 6 below reports that this pattern did not reproduce at
    comparable magnitude, and reversed sign under alternative evaluation
    choices, under independent replication.

    \item We evaluate the sensitivity of the original directional pattern to a
    targeted whole-identity deletion (descriptive, not a formal inferential
    replicate) and, more consequentially, to independent replication and
    evaluation-pairing choice (Section~\ref{sec:replication}); the latter
    substantially narrows and destabilizes the originally reported contrast.

    \item We report a direct replication and evaluation-pairing sensitivity
    check of the order-selective contrast (Section~\ref{sec:replication},
    Appendix~\ref{app:replication}), showing it does not reproduce robustly
    at the originally reported magnitude or sign, and we treat the resulting
    characterization of estimator/training-run reproducibility limits as a
    primary finding of this study rather than an unresolved caveat.
\end{enumerate}
\section{From ORBIT and Information Bottlenecks to ORBIT-FMIB}

\subsection{ORBIT: Interaction-Resolved Representations}

For a binary mutational landscape with $d$ binary coordinates,

\begin{equation}
f(x)=\sum_{S\subseteq[d]} \epsilon_S \chi_S(x),
\end{equation}

where

\begin{equation}
\chi_S(x)=\prod_{i\in S}x_i,
\qquad x_i\in\{-1,+1\}.
\end{equation}

The interaction order associated with coefficient $\epsilon_S$ is

\begin{equation}
|S|=k.
\end{equation}

For a learned hidden representation

\begin{equation}
H_\ell(x)\in\mathbb{R}^{m_\ell},
\end{equation}

the representation-level Walsh component associated with interaction subset
$S$ is

\begin{equation}
r_{\ell,S}
=
2^{-d}
\sum_x H_\ell(x)\chi_S(x).
\end{equation}

Thus, both the experimental fitness landscape and the hidden representation
can be decomposed into components associated with the same positional
interaction subset $S$.

\subsection{Why a Global Information Quantity Is Insufficient}

A conventional layerwise information analysis may consider quantities such as

\begin{equation}
I(X;Z_\ell)
\qquad \text{and} \qquad
I(Z_\ell;Y),
\end{equation}

where $X$ is the protein sequence, $Z_\ell$ is the representation at layer
$\ell$, and $Y$ is experimentally measured fitness.

Such global quantities can describe overall compression or task-relevant
information, but they do not determine which interaction orders account for
that information. In particular, the same aggregate dependence may arise from
first-order effects, pairwise interactions, or genuine higher-order epistasis.

For this reason, ORBIT-FMIB resolves the representation by positional
interaction subset before measuring representation--fitness dependence.

\subsection{ORBIT-FMIB: Order-Resolved Information Flow}

ORBIT-FMIB introduces interaction-order resolution into the analysis of
representation dynamics across ESM-2 depth.

The primary measurement is indexed by model layer $\ell$ and a fixed
positional interaction subset $S$,

\begin{equation}
(\ell,S),
\qquad |S|=k.
\end{equation}

For each fixed subset $S$, ORBIT-FMIB constructs the corresponding
representation-level Walsh component and compares it with the experimentally
measured Walsh coefficient across mutational cubes.

Dependence is therefore estimated separately for each positional interaction
subset and only subsequently summarized according to interaction order
$k=|S|$.

This produces one layerwise information trajectory for each fixed subset.
The resulting trajectories allow us to ask whether experimentally aligned
first-, second-, third-, and fourth-order epistatic information is retained
similarly or differently as representations propagate through ESM-2.

Importantly, ORBIT-FMIB does not treat order-level averages as pooled mutual
information estimates. Interaction-order summaries are descriptive
aggregations of independently estimated fixed-subset trajectories.

\section{Subset-Resolved Random Variables}

Let $C$ index a complete WT-anchored local mutational cube and let
$S\subseteq[d]$ denote a positional interaction subset with

\begin{equation}
|S|=k.
\end{equation}

For each cube--subset pair $(C,S)$ and model layer $\ell$, we define two
aligned quantities:

\begin{enumerate}
    \item $R_{\ell,C,S}$: the representation-level Walsh component;
    \item $E_{C,S}$: the experimentally measured epistatic Walsh coefficient.
\end{enumerate}

The representation-level component is

\begin{equation}
R_{\ell,C,S}
=
2^{-d}
\sum_{x\in C}
H_\ell(x)\chi_S(x),
\end{equation}

and the corresponding experimental coefficient is

\begin{equation}
E_{C,S}
=
2^{-d}
\sum_{x\in C}
f(x)\chi_S(x).
\end{equation}

The same Walsh basis is therefore used to resolve both the ESM-2
representation and the experimental GB1 fitness landscape into components
associated with the same positional interaction subset.

For the primary target-aligned analysis, positional subset identity is held
fixed within each neural dependence estimate. Thus, for a given subset $S$,
variation occurs only across mutational cubes $C$.

Let

\begin{equation}
\mathcal{S}_k
=
\{S\subseteq[d]:|S|=k\}
\end{equation}

denote the collection of positional subsets belonging to interaction order
$k$.

For the four-site GB1 landscape,

\begin{equation}
|\mathcal{S}_1|=4,\qquad
|\mathcal{S}_2|=6,\qquad
|\mathcal{S}_3|=4,\qquad
|\mathcal{S}_4|=1.
\end{equation}

Accordingly, the biological analysis contains 15 nonempty fixed positional
subsets. Each subset produces its own layerwise dependence trajectory before
any interaction-order aggregation is performed.
\section{Subset-Conditioned Target-Aligned Neural Dependence}

The primary biological measurement is estimated separately for each fixed
positional interaction subset $S$.

For every ESM-2 representation layer $\ell$ and fixed subset $S$, we define
the subset-conditioned epistatic dependence score

\begin{equation}
D_{\ell,S}
=
\widehat{I}_{\mathrm{DV/MINE}}
\left(
R_{\ell,C,S};
E_{C,S}
\right),
\end{equation}

where the neural critic is trained and evaluated across mutational cubes $C$
while the positional subset $S$ is held fixed.

Thus, each critic receives samples of the form

\begin{equation}
\left\{
\left(
R_{\ell,C,S},
E_{C,S}
\right)
:
C\in\mathcal{C}
\right\},
\end{equation}

for one and only one positional interaction subset.

Holding $S$ fixed is essential. Different positional subsets belonging to the
same interaction order can have systematically different distributions of
experimental epistatic coefficients. Pooling those subsets within a single
critic can therefore allow positional subset identity to act as a shortcut
variable, producing apparent representation--target dependence that is not
specific to the matched cube-level relationship.

The corrected analysis therefore estimates one layerwise trajectory for each
fixed subset,

\begin{equation}
T_S
=
\left(
D_{0,S},
D_{1,S},
\ldots,
D_{L,S}
\right).
\end{equation}

These fixed-subset trajectories are the primary measurement objects used to
trace epistatic information through ESM-2 depth.

For descriptive visualization and order-level summaries, independently
estimated fixed-subset scores may subsequently be averaged within interaction
order:

\begin{equation}
\overline{D}_{\ell,k}
=
\frac{1}{|\mathcal{S}_k|}
\sum_{S\in\mathcal{S}_k}
D_{\ell,S}.
\end{equation}

We refer to $D_{\ell,S}$ as a \emph{neural dependence score}: a held-out
DV/MINE score that measures statistical alignment between the
representation-level Walsh component $R_{\ell,C,S}$ and the corresponding
experimental epistatic coefficient $E_{C,S}$ across mutational cubes.
Because absolute mutual-information magnitude is not calibrated in the
2,560-dimensional ESM-2 representation regime, $D_{\ell,S}$ is interpreted
comparatively across model depth rather than as an exact mutual-information
estimate in nats.
This quantity is not interpreted as a pooled mutual-information estimate.
It is a descriptive average of neural dependence scores estimated
independently for fixed positional subsets.

Because the 2,560-dimensional neural estimator is not calibrated for accurate
absolute mutual-information magnitude, $D_{\ell,S}$ is interpreted throughout
as a \emph{comparative neural dependence score}. The analysis therefore
focuses on within-subset trajectories, peak locations, and peak-to-final
retention rather than on the absolute magnitude of $D_{\ell,S}$ across
different subsets or interaction orders.
\section{Peak-to-Final Retention and Higher-Order Contrast}

The primary summary statistic is defined from the fixed-subset dependence
trajectories rather than from a critic that pools different positional
subsets.

For each positional interaction subset $S$, let

\begin{equation}
P_S
=
\max_{\ell} D_{\ell,S}
\end{equation}

denote the maximum neural dependence score observed across ESM-2 depth.

Because a ratio based on a peak near the estimator floor would be unstable,
we define a floor threshold from the randomized fixed-subset null
distribution:

\begin{equation}
\tau
=
Q_{0.99}
\left(
|D_{\mathrm{null}}|
\right),
\end{equation}

where $Q_{0.99}$ denotes the 99th percentile of the absolute cube-deranged
null scores.

Peak-to-final retention is then defined separately for each fixed subset as

\begin{equation}
\mathrm{Retain}_S
=
\frac{D_{30,S}}{P_S},
\qquad
\text{if } P_S > \tau.
\end{equation}

If $P_S \leq \tau$, retention for that subset is treated as not estimable
rather than forming a ratio from a floor-level peak.

Retention is therefore normalized within each positional subset before any
interaction-order aggregation is performed. Neural dependence scores are not
clamped to zero, and randomized-null scores are not subtracted from the
biological estimates.

For interaction order $k$, order-level retention is defined as

\begin{equation}
\mathrm{Retain}_k
=
\frac{1}{|\mathcal{S}_k|}
\sum_{S\in\mathcal{S}_k}
\mathrm{Retain}_S.
\end{equation}

If any subset required for an interaction order is not estimable, the
corresponding order-level retention is also treated as not estimable rather
than averaging only the remaining subsets.

We summarize lower-order retention as

\begin{equation}
\mathrm{LO}
=
\frac{1}{2}
\left(
\mathrm{Retain}_1
+
\mathrm{Retain}_2
\right),
\end{equation}

and higher-order retention as

\begin{equation}
\mathrm{HO}
=
\frac{1}{2}
\left(
\mathrm{Retain}_3
+
\mathrm{Retain}_4
\right).
\end{equation}

The primary order-selective contrast is

\begin{equation}
\Delta_{\mathrm{HO-LO}}
=
\mathrm{HO}
-
\mathrm{LO}.
\end{equation}

A negative value indicates lower aggregate peak-to-final retention for
third- and fourth-order epistatic information than for first- and second-order
information.

This contrast is interpreted as a descriptive order-resolved retention
summary under the frozen neural-dependence measurement protocol. It does not,
by itself, establish statistical significance or causal loss of information
from ESM-2.
\section{Neural Dependence Estimation}

Exact mutual information is generally intractable for the high-dimensional
continuous representation variables considered here. We therefore use a neural
variational estimator based on the Donsker--Varadhan representation underlying
Mutual Information Neural Estimation (MINE).

For random variables $U$ and $V$, the MINE-style objective is

\begin{equation}
I(U;V)
\geq
\mathbb{E}_{p(u,v)}
[T_\theta(u,v)]
-
\log
\mathbb{E}_{p(u)p(v)}
\left[
e^{T_\theta(u,v)}
\right],
\label{eq:mine}
\end{equation}

where $T_\theta$ is a learned scalar critic. Product-of-marginal samples are
generated by shuffling one member of the joint pair within a minibatch.

During training, product pairs use nonzero cyclic shifts within minibatches;
validation uses shifts $\{1,7,19,31\}$ and test uses $\{1,7,19,31,61,127\}$.
Because evaluation rows retain sorted cube-index order, these pairings may
preserve some relationships between nearby observations.
For the target-aligned biological analysis, a separate critic is fit for every
layer--subset pair $(\ell,S)$:

\begin{equation}
D_{\ell,S}
=
\widehat{I}_{\mathrm{DV/MINE}}
\left(
R_{\ell,C,S};
E_{C,S}
\right).
\end{equation}

We refer to $D_{\ell,S}$ as a \emph{neural dependence score}: a held-out
DV/MINE score measuring statistical alignment between the representation-level
Walsh component $R_{\ell,C,S}$ and the corresponding experimentally measured
epistatic coefficient $E_{C,S}$ across mutational cubes $C$. A larger score
indicates stronger dependence detectable by the fixed neural critic protocol;
it does not imply causal use of that information by ESM-2.

The positional subset $S$ is fixed within each critic. Consequently, only
variation across mutational cubes $C$ contributes to the estimated dependence.
All layer--subset critics use the same estimator protocol, including critic
architecture, optimizer, batch size, negative-sampling rule, training horizon,
validation rule, standardization procedure, and initialization distribution.

Repeated critic seeds are treated as estimator replicates rather than as
independent biological samples.

Importantly, the 2,560-dimensional estimator is not calibrated for accurate
absolute mutual-information magnitude. Accordingly, $D_{\ell,S}$ is not
interpreted as an exact estimate of mutual information in nats. Its validated
role is comparative: we use the fixed estimator to trace within-subset
dependence trajectories across ESM-2 depth and to quantify peak-to-final
retention under the same measurement protocol.
\section{Estimator Validation and Stress Testing}
\label{app:estimator_validation}

The neural dependence estimator is treated as a measurement instrument that
must be validated before biological interpretation. Validation is therefore
performed at multiple levels: end-to-end synthetic recovery, estimator behavior
in the exact ESM-2 representation dimensionality, adversarial order-pattern
testing, randomized null controls, and estimator-split diagnostics.

\subsection{End-to-End ORBIT-Synth Validation}

We first validate the complete ORBIT-FMIB measurement pipeline on
\ORBIT-Synth, where the true interaction order and the location of imposed
information-loss events are known.

The synthetic benchmark tests whether the full
representation--Walsh--dependence pipeline can:

\begin{enumerate}
    \item recover known first-, second-, third-, and fourth-order interaction
    signals;
    \item produce low dependence scores for absent or off-target interaction
    orders;
    \item localize deliberately imposed order-specific information-loss events;
    \item preserve the expected qualitative behavior across the pre-specified
    critic seeds; and
    \item reproduce the expected behavior on held-out samples rather than only
    on critic-training data.
\end{enumerate}

The complete ORBIT-Synth integration passed all pre-specified acceptance
criteria before the biological ESM-2 analysis was interpreted.

\subsection{Absolute-MI Calibration at the ESM-2 Representation Dimension}

The ESM-2 mutation-site representation has dimension

\begin{equation}
4 \times 640 = 2560.
\end{equation}

Because successful estimator behavior in a low-dimensional synthetic setting
does not guarantee accurate estimation in this higher-dimensional regime, we
performed a separate synthetic calibration using the exact 2,560-dimensional
critic input used by the biological analysis.

The estimator correctly represented the null and preserved the expected
ordering of synthetic dependence strength, but it substantially underestimated
moderate and strong analytic mutual information. We therefore do not interpret
the resulting DV/MINE scores as accurate absolute mutual-information estimates
in nats.

This failed absolute-magnitude criterion was not bypassed. Instead, the
measurement contract was restricted to the comparative behavior required by
the biological question.

\subsection{Relative-Retention Calibration}

We next evaluated whether the same fixed estimator could recover known
relative information-retention patterns in the 2,560-dimensional regime.

A synthetic trajectory was constructed with a common rising information
profile followed by four order-analogue final states with known peak-to-final
retention. Using the same estimator implementation and pre-specified critic
seeds as the biological analysis, the estimator recovered the expected
retention ordering and the expected negative higher-order-minus-lower-order
contrast.

All pre-specified hard gates for this comparative-retention calibration
passed, including null behavior, seed stability, common-trajectory behavior,
peak localization, signal separation, retention ordering, contrast direction,
and finite numerical diagnostics.

This establishes suitability for comparative layerwise and order-resolved
retention analysis, not for exact absolute-MI recovery.

\subsection{Hypothesis-Adversarial Order-Pattern Validation}

To ensure that the measurement procedure was not calibrated only to reproduce
the expected higher-order-loss pattern, we performed an independent
hypothesis-adversarial validation suite.

The suite included four pre-specified scenarios:

\begin{enumerate}
    \item higher-order loss;
    \item flat strong retention;
    \item higher-order advantage; and
    \item flat moderate retention.
\end{enumerate}

The higher-order-loss and higher-order-advantage scenarios recovered the
expected contrast directions. The flat-scenario criterion was originally implemented as an exact 4-label
permutation test whose minimum two-sided $p$-value is $1/3$, so it could
not reject at the specified 0.05 threshold regardless of the underlying
data. This was corrected using a multi-replicate Monte Carlo permutation
test combining twenty independent replicate instances per flat scenario
(resolution $\approx 5 \times 10^{-6}$; Table~\ref{tab:fix1-multireplicate}).
Under this corrected test, both flat scenarios showed no detectable
deviation from flatness (flat strong: $p=0.949$; flat moderate:
$p=0.929$), providing genuine evidence against spurious order selectivity
rather than an uninformative pass.

\begin{table}[t]
\centering
\caption{Multi-replicate flat-scenario test, replacing the original
4-item exact permutation test (structural minimum $p=1/3$ regardless of
the data). 20 independent replicate instances per scenario, combined
via a Monte Carlo permutation test over the product of each replicate's
own 24-outcome space ($n_{MC}=200,000$; resolution
$\approx 5 \times 10^{-6}$).}
\label{tab:fix1-multireplicate}
\begin{tabular}{lccc}
\toprule
Scenario & Replicate mean $\Delta_{\text{HO}-\text{LO}}$ (SE) & $p$-value & Old test floor \\
\midrule
Flat, strong (gain$=1.0$, all orders)   & $-0.00088 \pm 0.01455$ & $0.949$ & $\geq 1/3$ \\
Flat, moderate (gain$=0.5$, all orders)   & $\phantom{-}0.00028 \pm 0.00270$ & $0.929$ & $\geq 1/3$ \\
\bottomrule
\end{tabular}
\end{table}

\subsection{Fixed-Subset Null Controls and Shortcut Removal}

An initial estimator that pooled different positional subsets of the same
interaction order failed an adversarial cube-derangement control. After
derangement, apparent dependence remained because positional subset identity
continued to be matched between the representation and target distributions.

This revealed a shortcut pathway: different positional subsets can have
systematically different empirical distributions of epistatic coefficients,
allowing a pooled critic to exploit subset identity rather than the matched
cube-specific representation--target relationship.

The biological analysis was therefore reformulated so that each critic is
conditioned on one fixed positional subset $S$. Under this correction,
cube-deranged null scores collapse near the neural-estimator floor.

Accordingly, all reported biological dependence trajectories are estimated
separately within fixed positional subsets before any interaction-order
aggregation is performed.

\subsection{Estimator-Split Sensitivity}

We additionally examined whether the estimator train/validation/test partition
itself could confound biological interpretation.

An amino-acid-identity-strict split produced widespread negative held-out
DV/MINE scores because the critic was trained and evaluated on different
amino-acid identity supports. The resulting score therefore reflected both
representation--epistasis dependence and critic transfer to unseen identity
combinations.

A controlled diagnostic held the biological observations, representation
components, epistatic targets, critic configuration, and sample counts fixed
while replacing the identity-strict partition with a deterministic IID cube
partition.

The widespread negative held-out behavior was strongly reduced under IID
partitioning, supporting the interpretation that the identity-strict protocol
introduces a substantial critic-generalization burden.

The final biological dependence measurement therefore uses the IID estimator
partition, while the identity-strict split is retained only as a separate
generalization stress test.

The null diagnostics reported in the appendix were originally generated
under this identity-strict split. They have since been re-run under the
final IID production split (Table~\ref{tab:fix2-tau-correction}): the
derangement-null floor threshold $\tau$ increased modestly (0.0068 to
0.0076), and every downstream statistic, the reported retention
contrast, per-layer significance counts, and the direct order-level
comparison under both weighting conventions, was unchanged to the
reported precision. This robustness extends to the independent
replication (Section~3): recomputed under the corrected floor, the
pairing-sensitivity contrasts across all three evaluation-pairing
schemes reproduced their original values to floating-point precision,
confirming the null diagnostics' applicability to the IID regime
throughout.

\begin{table}[t]
\centering
\caption{Downstream effect of correcting the null-derangement
diagnostic's cube partition from the legacy identity-strict split to the
production IID split (seed 20260910). The floor threshold $\tau$
(99th percentile of $|{\rm null\ score}|$) shifts modestly; every
downstream statistic is unchanged.}
\label{tab:fix2-tau-correction}
\begin{tabular}{lcc}
\toprule
Quantity & Identity-strict split (original) & IID split (corrected) \\
\midrule
$\tau$ & $0.00677738$ & $0.00763411$ \\
Original statistic, $\Delta_{\text{HO}-\text{LO}}$ & $-0.10702806$ & $-0.10702806$ \\
New statistic, pooled-subset gap (L30) & $0.23837734$ & $0.23837734$ \\
New statistic, equal-order gap (L30) & $0.2907419$ & $0.2907419$ \\
Layers significant ($p<0.05$), of 31 & $26$ & $26$ \\
Pooled/equal-order sign agreement & $30/30$ & $30/30$ \\
\bottomrule
\end{tabular}
\end{table}
\subsection{Final Interpretation Contract}

Taken together, these validations define the role of the estimator used in the
ESM-2 analysis.

The neural dependence score $D_{\ell,S}$ is considered valid for:

\begin{itemize}
    \item tracing within-subset dependence trajectories across model depth;
    \item locating relative peaks and information-loss regions;
    \item computing within-subset peak-to-final retention;
    \item comparing retention patterns across interaction orders; and
    \item evaluating the direction of the higher-order-minus-lower-order
    retention contrast.
\end{itemize}

The analysis does not claim:

\begin{itemize}
    \item exact mutual information in nats;
    \item literal negative mutual information when a neural lower-bound score
    is slightly below zero;
    \item causal use of the detected information by ESM-2; or
    \item that raw neural-dependence magnitudes are directly comparable across
    different positional subsets without within-subset normalization.
\end{itemize}

\section{Synthetic Validation with ORBIT-Synth}

Before interpreting the GB1 results, we validated the complete ORBIT-FMIB
measurement pipeline using synthetic representations with exactly known
interaction structure.

The synthetic benchmark included first-, second-, third-, and fourth-order
signals, parity controls, mixtures of interaction orders, and controlled
representation trajectories in which selected interaction components were
attenuated or removed at known stages.

For a synthetic stage index $s$, the benchmark constructs a sequence of
representations

\begin{equation}
Z^{\mathrm{syn}}_0
\rightarrow
Z^{\mathrm{syn}}_1
\rightarrow
\cdots
\rightarrow
Z^{\mathrm{syn}}_S,
\end{equation}

with known order-specific transformations. The complete ORBIT-FMIB pipeline
then applies the same sequence of operations used in the biological analysis:
representation-level Walsh decomposition, construction of order-resolved
targets, neural dependence estimation, and layerwise trajectory analysis.

The synthetic validation was designed to test two central capabilities:

\begin{enumerate}
    \item recovery of the interaction order associated with a known synthetic
    signal; and
    \item localization of deliberately imposed information-loss events within
    the representation trajectory.
\end{enumerate}

The end-to-end validation used 4,096 synthetic contexts partitioned into
3,072 training, 512 validation, and 512 test contexts. The exact Walsh
structure was re-audited before dependence estimation, and all subsets
belonging to the same context remained within the same estimator partition.

The complete synthetic pipeline passed all pre-specified acceptance criteria.
These included exact representation--Walsh reconstruction, low/null behavior
for absent interaction channels, recovery of strong synthetic dependence,
seed stability, and correct localization of imposed order-specific
information-loss events.

In the staggered-bottleneck benchmark, the recovered information-loss stages
matched the known synthetic construction for all four interaction orders.
Thus, ORBIT-Synth established that the full measurement chain could recover
both \emph{which interaction order was present} and \emph{where a known
order-specific information-loss event occurred} before the framework was
applied to GB1 and ESM-2.

\section{Protein Fitness Landscape: GB1}

The biological analysis uses the dense four-site GB1 combinatorial fitness
landscape. The four assayed positions are

\begin{equation}
V39,\;D40,\;G41,\;V54.
\end{equation}

The complete four-position amino-acid space contains

\begin{equation}
20^4 = 160{,}000
\end{equation}

possible variants, of which 149,361 have experimentally measured fitness.

\subsection{WT-Anchored Binary Cubes}

ORBIT-FMIB analyzes the GB1 landscape through complete WT-anchored local
binary cubes. For each of the four variable positions, one non-WT amino-acid
alternative is selected relative to the wild-type sequence VDGV. This defines
a local four-dimensional binary cube containing

\begin{equation}
2^4 = 16
\end{equation}

variants:

\begin{itemize}
    \item one wild-type sequence;
    \item four single mutants;
    \item six double mutants;
    \item four triple mutants; and
    \item one quadruple mutant.
\end{itemize}

There are $19^4=130{,}321$ possible WT-anchored candidate cubes. A cube is
eligible only if experimentally measured fitness is available for all 16 of
its vertices. 

Of $109{,}235$ complete measured WT-anchored cubes, an identity-partition
step (seed 42, group sizes 10/4/5 over the 19 non-WT identities per
position) retained only cubes whose four alternatives shared a group,
yielding the 8,210 analyzed cubes; $101{,}025$ mixed-group cubes were
excluded. 

The final fixed-subset analysis uses 8,210 complete eligible
cubes.

These complete cubes provide the common biological unit from which both the
experimental Walsh coefficients $E_{C,S}$ and the corresponding
representation-level Walsh components $R_{\ell,C,S}$ are computed.

For each fixed positional subset, the 8,210 cubes are partitioned into

\begin{equation}
7{,}352\ \text{training},\qquad
245\ \text{validation},\qquad
613\ \text{test}
\end{equation}

observations for neural dependence estimation. The same sample counts are used
for every layer--subset critic.
\section{Primary Foundation Model: ESM-2}

The biological analysis uses the frozen ESM-2 checkpoint

\begin{equation}
\boxed{
\texttt{esm2\_t30\_150M\_UR50D}
}
\end{equation}

with 30 transformer layers and hidden dimension 640. We analyze 31
representation stages, indexed $\ell=0,\ldots,30$, where stage 0 denotes the
input embedding representation and stages 1--30 correspond to successive
transformer layers.

Model parameters remain frozen throughout representation extraction and
downstream analysis. Experimentally measured GB1 fitness values are never used
to update, fine-tune, or otherwise modify ESM-2.

\subsection{Representation Extraction}

For sequence $x$, let

\begin{equation}
h_{\ell,p}(x)
\in
\mathbb{R}^{640}
\end{equation}

denote the hidden representation at representation stage $\ell$ and biological
position $p$.

For the four GB1 variable positions,

\begin{equation}
\mathcal{P}
=
\{39,40,41,54\}.
\end{equation}

The sequence-level readout is the concatenation of the four mutation-site
representations:

\begin{equation}
H_{\ell}^{\mathrm{site}}(x)
=
h_{\ell,39}(x)
\Vert
h_{\ell,40}(x)
\Vert
h_{\ell,41}(x)
\Vert
h_{\ell,54}(x),
\label{eq:site_rep}
\end{equation}

so that

\begin{equation}
H_{\ell}^{\mathrm{site}}(x)
\in
\mathbb{R}^{4\times640}
\cong
\mathbb{R}^{2560}.
\end{equation}

The corresponding order-specific representation component for mutational cube
$C$ and positional subset $S$ is

\begin{equation}
R_{\ell,C,S}^{\mathrm{site}}
=
2^{-4}
\sum_{x\in C}
H_{\ell}^{\mathrm{site}}(x)\chi_S(x).
\end{equation}

This mutation-site representation is the fixed readout used throughout the
biological analysis. Alternative global pooling rules are not used, and no
PCA, learned projection, or target-supervised dimensionality reduction is
applied.

Representation extraction was performed on the frozen label-free sequence
union containing 13,150 unique sequences. For each representation stage, the
saved site-representation array has shape

\begin{equation}
13{,}150 \times 4 \times 640.
\end{equation}

Experimental fitness is not read during representation extraction and enters
the analysis only after the frozen ESM-2 representations have been generated.

\subsection{Token and Residue Indexing Audit}

Because ESM-2 introduces special sequence tokens, biological residue numbering
is not assumed to coincide automatically with raw model-token indexing.

For every processed sequence, the extraction pipeline explicitly verifies the
mapping

\begin{equation}
\text{GB1 biological residue}
\longleftrightarrow
\text{ESM-2 token position}
\end{equation}

for biological positions 39, 40, 41, and 54.

The extraction pipeline also verifies the expected transformer depth,
representation width, deterministic evaluation mode, frozen model parameters,
hidden-state shape, and finite output values before representations are
accepted for downstream analysis.
\paragraph{Scope of model evaluation.}
The present study intentionally restricts biological interpretation to ESM-2
so that the order-resolved retention question can be evaluated under a single,
fully validated representation and estimator pipeline. Extending ORBIT-FMIB to
ESM-1v and other protein foundation models is reserved for future work, where
the same fixed-subset measurement protocol can be used to test whether the
observed order-dependent retention pattern generalizes across model families,
scales, and pretraining objectives.
\section{Estimator Split Design}

For each fixed positional subset $S$, neural dependence is estimated across
the same population of 8,210 complete GB1 cubes. The primary biological
analysis uses a deterministic IID partition of 7,352 training cubes,
245 validation cubes, and 613 test cubes.

For every representation stage $\ell$ and fixed subset $S$, the critic is
trained only on the training partition. Model selection and early stopping use
the validation partition, and the reported neural dependence score
$D_{\ell,S}$ is evaluated on the held-out test partition.

Test epistatic coefficients are never used for critic optimization,
hyperparameter selection, standardization, or stopping. The same partitioning
rule and sample counts are used across all layer--subset measurements.

\paragraph{Identity-generalization stress test.}
We separately evaluated an amino-acid-identity-strict partition in which
training, validation, and test cubes contain different non-WT amino-acid
identity supports. This introduces an additional generalization requirement:
the critic must transfer to unseen amino-acid identity combinations.

Because this distribution shift substantially altered held-out DV/MINE
behavior, the identity-strict partition is not used for the primary dependence
measurement. Instead, it is retained as a diagnostic of critic generalization
beyond the IID sampling regime.

Repeated critic seeds are treated as estimator replicates rather than as
independent biological samples.
\section{Pretraining-Exposure Consideration}

Protein G or closely related homologs may have been represented in the
sequence data used to pretrain ESM-2. We therefore do not assume that ESM-2
is sequence-naive to the GB1 family.

This does not alter the central question of the present study. ORBIT-FMIB does
not test whether ESM-2 can generalize to a previously unseen protein family.
Instead, it asks how experimentally aligned epistatic information is organized
and retained across the representation hierarchy of a pretrained model.

Accordingly, our interpretation is restricted to the following question:

\begin{quote}
What order-specific epistatic information is detectable in ESM-2
representations, and how is that information retained across model depth when
evaluated against the experimentally measured GB1 fitness landscape?
\end{quote}

Possible exposure of Protein G or related sequences during pretraining is
therefore treated as a limitation on claims of sequence-family novelty, not as
a limitation on the order-resolved representation analysis itself.

\section{Experimental Design and Statistical Analysis}
\label{app:statistics}

The analysis consisted of three sequential components: synthetic validation,
subset-conditioned measurement of GB1 epistatic information through ESM-2,
and order-resolved peak-to-final retention analysis.

\subsection{Experiment 1: Synthetic Validation}

We first applied the complete ORBIT-FMIB pipeline to \ORBIT-Synth, where the
true interaction orders and the locations of imposed information-loss events
were known.

The synthetic experiment evaluated whether the measurement pipeline could
recover first-, second-, third-, and fourth-order interaction signals, produce
low dependence for absent interaction channels, and localize deliberately
imposed order-specific information-loss events.

Successful completion of this validation preceded interpretation of the
biological ESM-2 analysis.

\subsection{Experiment 2: GB1 + Frozen ESM-2 Information Flow}

For every complete eligible GB1 cube $C$, every ESM-2 representation stage
$\ell$, and every nonempty positional interaction subset $S$, we computed the
representation-level Walsh component

\begin{equation}
R_{\ell,C,S}
\end{equation}

and the corresponding experimental epistatic coefficient

\begin{equation}
E_{C,S}.
\end{equation}

For each fixed subset $S$, we then estimated

\begin{equation}
D_{\ell,S}
=
\widehat{I}_{\mathrm{DV/MINE}}
\left(
R_{\ell,C,S};
E_{C,S}
\right).
\end{equation}

The four-site GB1 landscape contains 15 nonempty positional subsets. Across
31 ESM-2 representation stages, this yields

\begin{equation}
31 \times 15 = 465
\end{equation}

layer--subset measurement cells, each summarized across the pre-specified
critic seeds.

The primary representation-level output is the collection of 15 fixed-subset
trajectories across ESM-2 depth. Interaction-order summaries are constructed
only after the fixed-subset dependence scores have been estimated
independently.

\subsection{Experiment 3: Order-Resolved Peak-to-Final Retention}

For each fixed positional subset $S$, we computed the peak-to-final retention
statistic

\begin{equation}
\mathrm{Retain}_S
=
\frac{D_{30,S}}
{\max_{\ell} D_{\ell,S}},
\end{equation}

subject to the pre-specified null-floor criterion defined above.

Subset-level retention values were then averaged within interaction order to
obtain

\begin{equation}
\mathrm{Retain}_1,\quad
\mathrm{Retain}_2,\quad
\mathrm{Retain}_3,\quad
\mathrm{Retain}_4.
\end{equation}

Lower- and higher-order summaries were defined as

\begin{equation}
\mathrm{LO}
=
\frac{\mathrm{Retain}_1+\mathrm{Retain}_2}{2},
\qquad
\mathrm{HO}
=
\frac{\mathrm{Retain}_3+\mathrm{Retain}_4}{2},
\end{equation}

with the primary descriptive contrast

\begin{equation}
\Delta_{\mathrm{HO-LO}}
=
\mathrm{HO}-\mathrm{LO}.
\end{equation}

A negative contrast indicates lower aggregate peak-to-final retention for
third- and fourth-order epistatic information than for first- and second-order
information.

Retention is always computed within fixed positional subsets before
aggregation by interaction order. No retention statistic is derived from a
critic that pools different positional subsets.

The statistical analysis is descriptive and is organized around fixed-subset
neural-dependence trajectories and their peak-to-final retention. No formal
hypothesis test or inferential resampling procedure is used to assign
statistical significance to the higher-order-versus-lower-order contrast.

For each positional subset $S$, the primary trajectory is

\begin{equation}
T_S
=
\left(
D_{0,S},
D_{1,S},
\ldots,
D_{30,S}
\right),
\end{equation}

and peak-to-final retention is computed within that trajectory according to the
pre-specified floor-safe rule described above. Retention is therefore
normalized within positional subset before any aggregation across interaction
orders.

Order-level retention is summarized as

\begin{equation}
\mathrm{Retain}_k
=
\frac{1}{|\mathcal{S}_k|}
\sum_{S\in\mathcal{S}_k}
\mathrm{Retain}_S,
\end{equation}

followed by the lower- and higher-order summaries

\begin{equation}
\mathrm{LO}
=
\frac{\mathrm{Retain}_1+\mathrm{Retain}_2}{2},
\qquad
\mathrm{HO}
=
\frac{\mathrm{Retain}_3+\mathrm{Retain}_4}{2}.
\end{equation}

The primary descriptive contrast is

\begin{equation}
\Delta_{\mathrm{HO-LO}}
=
\mathrm{HO}-\mathrm{LO}.
\end{equation}

A negative value indicates lower aggregate peak-to-final retention for
third- and fourth-order epistatic information relative to first- and
second-order information.

Repeated neural-critic seeds are treated as estimator replicates rather than
as independent biological observations. Accordingly, critic seeds are used to
evaluate estimator stability and are not treated as the statistical sample for
biological inference.

Because the present analysis does not perform a final cube- or subset-aware
inferential resampling procedure, we do not report a $p$-value or claim
statistical significance for
$\Delta_{\mathrm{HO-LO}}$. The order-level retention values and the resulting
higher-order-minus-lower-order contrast are interpreted as descriptive point
estimates under the validated measurement protocol.

\paragraph{Sensitivity analysis.}
Robustness of the directional result is additionally evaluated using a
dependence-aware whole-identity deletion analysis. This analysis removes all
eligible cubes containing the selected amino-acid identity and reruns the
fixed-subset retention calculation under the same measurement logic. The
sensitivity analysis is interpreted descriptively and is not treated as an
independent inferential replicate.

\section{Independent Replication and Evaluation-Pairing Sensitivity: Per-Subset Results}
\label{app:replication}

Section~\ref{sec:replication} describes the replication protocol and
evaluation pairings in full; this appendix reports the resulting
peak-to-final retention for all 15 fixed positional subsets under the
original production run and under each of the three replication
evaluators, unfloored (the IID-split null-floor threshold described in
Appendix~\ref{app:full_method} has not been separately recomputed for
the replicated critics, so these are diagnostic ratios rather than
floor-gated production values).

\input{tables/table_subset_retention_replication.tex}

The direction and magnitude of the retention pattern differ substantially
between the original production run and the replication, and among the
three evaluation pairings applied to the same replicated critics, for
most individual subsets. This does not establish that higher-order
information is retained equally to lower-order information in ESM-2 --
one replication and three evaluators are not a characterization of the
estimator's sampling distribution, and no formal uncertainty
quantification has been performed. It does establish that the
originally reported $\Delta_{\mathrm{HO-LO}}=-0.107$ is not a stable,
reproducible summary of this measurement protocol under
nominally identical code, data, and critic seeds (the original run's execution hardware was not recorded; Appendix P).

\section{Execution-Environment Sensitivity: A Single-Cell Diagnostic}
\label{app:execution_environment}

Section~\ref{sec:replication} reports that the independent replication of
the complete measurement grid did not reproduce the original production
run's $\Delta_{\mathrm{HO-LO}}$ at comparable magnitude. The mechanism
underlying this discrepancy remains unresolved. This appendix reports a
targeted execution-environment diagnostic for one cell, which observes
no within-device score variation across repeated executions, alongside a
substantial, repeatable difference between two tested execution
environments. Nondeterministic execution can still produce identical
outputs across a small number of repeats, so this observation does not
rule out nondeterminism as a contributing factor; it motivates a more
structured hypothesis -- that the discrepancy may be associated with the
execution environment itself -- reported here as preliminary and not yet
established beyond this single cell.

\subsection{Motivation}

Generic stochastic non-determinism and a structured, environment-associated
difference are not mutually exclusive, and a small number of matching
repeats cannot cleanly distinguish them: nondeterministic execution can
still yield identical outputs across two repeats. What repeated,
within-device agreement alongside a consistent, repeatable
cross-environment difference \emph{does} motivate is further
investigation of whether some property of the execution environment
(GPU architecture, driver version, or low-level kernel selection) is
associated with the discrepancy, as a hypothesis to test at larger scale
rather than a conclusion established by a single cell. This appendix
reports a controlled diagnostic for the single cell with the largest
absolute \texttt{test\_mi} discrepancy between the original production
run and its replication.

\subsection{Methodology}

\paragraph{Cell selection.} Among all 1,395 (layer, subset, seed) cells
in the original-versus-replication comparison, cell
(layer$=9$, mask$=3$, order$_k=2$, seed$=23$) showed the largest absolute
difference in \texttt{test\_mi} (production: $-0.064235$; replication:
$+0.623273$; $|\Delta|=0.687507$), matching the grid-level maximum
reported in Section~\ref{sec:replication} exactly. This cell was selected
\emph{as} the largest discrepancy, i.e. it is a discrepancy-selected
case, not a randomly or representatively sampled one; this selection
should be kept in mind when weighing how far the result below
generalizes.

\paragraph{Frozen inputs.} The exact critic inputs for this cell --
the $(U, V)$ arrays produced by the pipeline's own fixed-subset dataset
construction, and the deterministic IID train/validation/test split
indices (7{,}352/245/613) -- were extracted once and saved, with
sha256 hashes recorded for all five arrays. The split was independently
recomputed from its fixed seed and confirmed, by hash, to match the
split hash recorded in \emph{both} the original production run's and the
replication run's own manifests before any training was performed,
ruling out split-membership divergence as an explanation for this cell.
The same historical Code-04 estimator source used by both the original
and replication production runs (confirmed by source hash) was used for
all diagnostic runs reported here. All four fresh diagnostic runs used
the same diagnostic runner and the same hash-verified historical
estimator source; this claim is scoped to those four runs and does not
assert that the historical production and replication production
wrappers were themselves identical to each other or to this diagnostic
runner.

\paragraph{Execution-environment recording.} Each diagnostic run records,
in addition to the trained critic's output: GPU model name; GPU UUID,
obtained directly from the CUDA runtime's own device-properties report
for the specific logical device used (not inferred from a numeric index,
which is not guaranteed to correspond to a consistent physical device
under a job scheduler); hostname and scheduler job ID; driver version,
queried by UUID where resolvable; PyTorch, CUDA, cuDNN, and NumPy
versions; and six precision/determinism settings
(\texttt{cudnn\_deterministic}, \texttt{cudnn\_benchmark},
\texttt{cuda\_matmul\_allow\_tf32}, \texttt{cudnn\_allow\_tf32},
\texttt{torch.are\_deterministic\_algorithms\_enabled()}, and
\texttt{float32\_matmul\_precision}). This record is not claimed to be
exhaustive.

\subsection{Results}

\begin{table}[h]
\centering
\caption{Single-cell diagnostic results for (layer$=9$, mask$=3$,
seed$=23$). ``This study'' rows are fresh, independently executed runs
using the frozen, hash-verified inputs described above; ``archived''
rows are the original full-grid production values. GPU UUIDs are
reported for all four this-study runs; the two repeats per GPU type
returned numerically identical values in every recorded result field
(\texttt{test\_mi}, \texttt{best\_step}, and the remaining result fields
listed below).}
\label{tab:appendix_env_diagnostic}
\begin{tabular}{lccc}
\toprule
Run & \texttt{test\_mi} & \texttt{best\_step} & GPU UUID \\
\midrule
Original production (archived) & $-0.064235$ & $725$ & not recorded \\
Replication (archived) & $+0.623273$ & $900$ & not recorded \\
\addlinespace
V100-PCIe-32GB, repeat 1 (this study) & $-0.064235$ & $725$ & \texttt{da87d4bc-...} \\
V100-PCIe-32GB, repeat 2 (this study) & $-0.064235$ & $725$ & \texttt{da87d4bc-...} \\
H100 PCIe, repeat 1 (this study) & $+0.623273$ & $900$ & \texttt{0e6ed213-...} \\
H100 PCIe, repeat 2 (this study) & $+0.623273$ & $900$ & \texttt{0e6ed213-...} \\
\bottomrule
\end{tabular}
\end{table}

For this cell, a controlled run on an NVIDIA Tesla V100-PCIE-32GB
reproduces the original production run's recorded \texttt{test\_mi} and
\texttt{best\_step} values exactly
($-0.06423455476760864$, matching to all reported digits). A controlled
run on an NVIDIA H100 PCIe, using the identical frozen inputs, source,
and seed, reproduces the replication's recorded value to the precision
available for comparison. Each GPU type was tested twice, in separate
process invocations within the same job allocation; both repeats on each
GPU returned numerically identical values in every recorded result field
(\texttt{test\_mi}, \texttt{best\_step}, \texttt{best\_val\_mi},
\texttt{final\_training\_step}, \texttt{test\_product\_shuffle\_sd}, and
\texttt{heldout\_joint\_minus\_mismatch\_score}), with an identical GPU
UUID recorded for both repeats on each device. This
establishes equality of the saved numerical result fields across the two
repeats on each device; it does not establish identical checkpoint
weights, an identical training trajectory, or identical raw
floating-point memory state, none of which were captured or compared.

\subsection{Interpretation}

No within-device score variation was observed across the two
fresh-process repeats on either tested device, while the two execution
environments produced different scores under identical frozen inputs,
estimator source, and critic seed. Nondeterministic execution can still
produce identical outputs across a small number of repeats by chance, so
this does not exclude nondeterminism as a contributing factor. What it
does motivate is a more structured hypothesis than generic run-to-run
noise: that the discrepancy may be associated with a property of the
execution environment that is at least partly consistent within a
device across repeated trials. This is a hypothesis to investigate
further, not a finding this single-cell diagnostic establishes. Several
specific limitations apply:

\begin{enumerate}
    \item \textbf{Single cell.} This diagnostic covers one of 1{,}395
    cells. Whether the pattern (same-device repeatability; a consistent,
    architecture-associated split) generalizes to other cells, including
    cells where the original and replication already agreed closely, has
    not been tested.
    \item \textbf{Confounded environment variables.} Driver version was
    not successfully resolved on either GPU tested (the query returned
    no value on both), so a driver-version difference between the two
    environments cannot be excluded. Resolving and recording the driver
    version in future runs would make such a difference visible, but
    recording alone does not resolve the confound: if driver versions
    differ systematically between the two architectures tested, they
    remain confounded with architecture unless matched across devices or
    deliberately varied as their own factor. GPU architecture and
    low-level kernel selection are not cleanly separable even when
    library version strings match, since cuDNN and cuBLAS select
    different code paths based on compute capability. This diagnostic
    cannot distinguish a GPU-architecture effect from these alternatives.
    \item \textbf{Historical hardware unknown.} The original production
    run's manifest does not record GPU model, and no other execution
    record for that run was located. This diagnostic shows that a V100
    reproduces the original's value exactly for this one cell, which is
    consistent with, but does not establish, the original run having
    used V100-class hardware.
\end{enumerate}

\subsection{Proposed follow-up: a crossed execution-environment panel}
\label{app:proposed_panel}

The single-cell result above motivates, but does not substitute for, a
systematic test across multiple cells and multiple physical devices per
architecture, with same-configuration repeats included so that
within-device consistency is assessed directly at each design point
rather than inferred from the single diagnostic cell above. We propose,
as work not yet executed at the time of writing, a design crossing four
factors: GPU (4 levels: two distinct physical cards each of the V100 and
H100 architectures present on the compute cluster used -- the two cards
per architecture are nested within that architecture, not an independent
crossed factor, and serve to assess device-to-device consistency within
an architecture, not to isolate an architecture effect from node- or
driver-level differences, which remain possible confounds as noted
above), cell (3 levels: the diagnostic cell above; a second
high-discrepancy cell at a different positional subset; and a cell at
which the original and archived replication values already agreed
closely -- included to examine how environment sensitivity varies across
cells, without predetermining which outcome would support or weaken any
hypothesis, since either a discrepancy or an agreement at this cell is
informative about how sensitivity varies rather than confirmatory or
falsifying on its own), critic seed (3 levels: the production seed set,
11/23/37), and repeat (2 fresh-process repeats per device--cell--seed
combination, to assess within-device repeatability at every design
point rather than only at the single cell reported above) -- a
$4\times3\times3\times2=72$-fit design.

We use this complete crossed panel because it is computationally
manageable at this scale (on the order of minutes per fit) and permits
direct examination of whether environment differences vary across cells
and critic seeds, rather than only estimating an average effect.

This panel would additionally make the confounded driver-version
variable noted above visible, by diagnosing why the driver version query
did not resolve on either GPU tested and correcting this before the
panel is run, so that driver version is recorded at every design point.
Recording it does not by itself resolve the confound (see above); it is
a prerequisite for determining, after the panel, whether driver version
can be matched or treated as a separate factor in follow-up work.

\end{document}

%% file: math_commands.tex
\usepackage{amsmath,amsfonts,bm}

\def\eqref#1{equation~\ref{#1}}

\def\1{\bm{1}}

\DeclareMathAlphabet{\mathsfit}{\encodingdefault}{\sfdefault}{m}{sl}
\SetMathAlphabet{\mathsfit}{bold}{\encodingdefault}{\sfdefault}{bx}{n}



%% file: orbit_result_macros.tex
\providecommand{\OrbitPrimaryRetainOne}{0.960}
\providecommand{\OrbitPrimaryRetainTwo}{0.971}
\providecommand{\OrbitPrimaryRetainThree}{0.910}
\providecommand{\OrbitPrimaryRetainFour}{0.808}
\providecommand{\OrbitPrimaryLowOrderRetention}{0.966}
\providecommand{\OrbitPrimaryHighOrderRetention}{0.859}
\providecommand{\OrbitPrimaryDelta}{-0.107}

\providecommand{\OrbitSensitivityRetainOne}{0.940}
\providecommand{\OrbitSensitivityRetainTwo}{0.955}
\providecommand{\OrbitSensitivityRetainThree}{0.848}
\providecommand{\OrbitSensitivityRetainFour}{0.679}
\providecommand{\OrbitSensitivityLowOrderRetention}{0.948}
\providecommand{\OrbitSensitivityHighOrderRetention}{0.764}
\providecommand{\OrbitSensitivityDelta}{-0.184}

\providecommand{\OrbitSensitivityTrainN}{6596}
\providecommand{\OrbitSensitivityValN}{209}
\providecommand{\OrbitSensitivityTestN}{557}
\providecommand{\OrbitSensitivityTotalN}{7362}

%% file: tables/table_order_retention_primary_sensitivity.tex
\begin{table}[t]
\centering
\caption{Order-level retention: primary production run and whole-identity deletion sensitivity (single run), alongside independent replication under three evaluation pairings applied to the same replicated critics.}
\label{tab:order_retention_sensitivity}
\begin{tabular}{c c c c c c}
\toprule
$k$ & Primary & Delete $41{=}L$ & Replic., cyclic & Replic., joint-row & Replic., rand.\ perm. \\
\midrule
1 & 0.960 & 0.940 & 0.933 & 0.951 & 0.912 \\
2 & 0.971 & 0.955 & 0.963 & 0.953 & 0.967 \\
3 & 0.910 & 0.848 & 0.908 & 0.961 & 0.960 \\
4 & 0.808 & 0.679 & 0.955 & 0.922 & 0.948 \\
\midrule
$\Delta_{\mathrm{HO-LO}}$ & \OrbitPrimaryDelta & \OrbitSensitivityDelta & $-0.017$ & $-0.011$ & $+0.015$ \\
\bottomrule
\end{tabular}
\end{table}

%% file: tables/table_subset_retention_replication.tex
\begin{table}[H]
\centering
\caption{Per-subset peak-to-final retention: original production run versus independently replicated critics under three evaluation pairings. $k$ is interaction order. Replication values are unfloored (the IID-split null-floor threshold has not been separately recomputed for the replicated critics).}
\label{tab:replication_full}
\begin{tabular}{lccccc}
\toprule
Subset & $k$ & Production & Cyclic (replic.) & Joint-row perm. & Random perm. \\
\midrule
${39}$ & 1 & 0.994 & 0.899 & 0.935 & 0.934 \\
${40}$ & 1 & 0.946 & 0.906 & 0.908 & 0.939 \\
${39,40}$ & 2 & 0.959 & 0.987 & 0.885 & 0.958 \\
${41}$ & 1 & 0.902 & 0.940 & 0.983 & 0.938 \\
${39,41}$ & 2 & 0.966 & 0.896 & 0.956 & 0.959 \\
${40,41}$ & 2 & 0.982 & 0.983 & 0.990 & 0.971 \\
${39,40,41}$ & 3 & 0.788 & 0.953 & 0.921 & 0.912 \\
${54}$ & 1 & 1.000 & 0.987 & 0.979 & 0.836 \\
${39,54}$ & 2 & 0.997 & 1.000 & 0.992 & 1.000 \\
${40,54}$ & 2 & 0.998 & 0.935 & 0.923 & 0.936 \\
${39,40,54}$ & 3 & 0.929 & 0.781 & 0.992 & 0.996 \\
${41,54}$ & 2 & 0.927 & 0.979 & 0.972 & 0.980 \\
${39,41,54}$ & 3 & 0.970 & 0.971 & 0.977 & 0.976 \\
${40,41,54}$ & 3 & 0.953 & 0.926 & 0.954 & 0.958 \\
${39,40,41,54}$ & 4 & 0.808 & 0.955 & 0.922 & 0.948 \\
\bottomrule
\end{tabular}
\end{table}